\documentclass[letter, 10pt, conference]{ieeeconf}      % Use this line for a4 paper

\IEEEoverridecommandlockouts                              % This command is only needed if 
\usepackage{graphics} % for pdf, bitmapped graphics files
\usepackage{epsfig} % for postscript graphics files
\usepackage{mathptmx} % assumes new font selection scheme installed
\usepackage{times} % assumes new font selection scheme installed
\usepackage{amsmath} % assumes amsmath package installed
\usepackage{amssymb}  % assumes amsmath package installed
\usepackage{xcolor}

\usepackage[per-mode=fraction]{siunitx}
\usepackage{comment}
\usepackage[normalem]{ulem}
\usepackage{xspace}
\usepackage{xcolor}
\usepackage{lastpage}
\usepackage{fancyhdr}

\DeclareSIUnit\px{px}
\DeclareSIUnit\fps{fps}

\definecolor{OliveGreen}{RGB}{0,200,25}
\newcommand{\red}[1]{\textcolor{red}{#1}}

\newcommand{\darkgreen}[1]{\textcolor{OliveGreen}{#1}}

\newcommand{\ie}{i.\,e.,\xspace}

\newcommand{\ackhht}{The authors are with the High Performance Humanoid Technologies Lab, Institute for Anthropomatics and Robotics, Karlsruhe Institute of Technology (KIT), Germany. }

\newcommand{\ackJuBot}{This work has been supported by the Carl Zeiss Foundation through the JuBot project.}

\newif\iffinal

\newcommand{\replaced}[2]{%
	\iffinal%
	#2%
	\else%
	\red{\ifmmode\text{\sout{\ensuremath{#1}}}\else\sout{#1}\fi}\darkgreen{#2}%
	\fi%
}
\newcommand{\removed}[1]{%
	\iffinal%
	\else%
	\red{\ifmmode\text{\sout{\ensuremath{#1}}}\else\sout{#1}\fi}%
	\fi%
}

\usepackage[capitalise]{cleveref}
\usepackage{subcaption}
\usepackage{import}
\usepackage{stfloats}
\usepackage[sort]{cite}
\usepackage{lipsum}
\usepackage{fancyhdr}

\graphicspath{{figures/}}
\newcommand*{\img}[1]{%
    \raisebox{-.02\baselineskip}{%
        \includegraphics[
        height=1em,
        keepaspectratio,
        ]{#1}%
    }%
}

\usepackage[per-mode=fraction]{siunitx}
\usepackage{xfrac}
\usepackage{nicefrac}
\DeclareSIUnit{\Nm}{\newton\metre}
\DeclareSIUnit{\ms}{\milli\second}
\DeclareSIUnit{\mmps}{\milli\metre\per\second}
\DeclareSIUnit{\mps}{\metre\per\second}
\DeclareSIUnit{\MPa}{\mega\pascal}
\DeclareSIUnit{\hPa}{\hecto\pascal}
\DeclareSIUnit{\kPa}{\kilo\pascal}
\DeclareSIUnit{\kg}{\kilo\gram}

\usepackage[acronym]{glossaries}
\setacronymstyle{long-short}
\newacronym{RoM}{RoM}{range of motion}
\newacronym{RMSE}{RMSE}{root-mean-squared error}
\newacronym{DoF}{DoF}{degrees of freedom}
\newacronym{EMG}{EMG}{electromyography}
\newacronym{IMU}{IMU}{inertial measurement unit}
\newacronym{ADL}{ADL}{activities of daily living}
\newacronym{FMG}{FMG}{force myography}
\newacronym{FSR}{FSR}{force sensing resistor}
\newacronym{SMP}{SMP}{surface muscle pressure}
\newacronym{PFDF}{PF/DF}{plantar-/dorsiflexion}
\newacronym{INEV}{IN/EV}{in-/eversion}
\newacronym{IRER}{IR/ER}{internal/external rotation}
\newacronym{CPU}{CPU}{computer processing unit}
\newacronym{ME}{ME}{mean error}
\newacronym{GM}{GM}{gastrocnemius medialis}
\newacronym{GL}{GL}{gastrocnemius lateralis}
\newacronym{TA}{TA}{tibialis anterior}
\newacronym{PAFO}{PAFO}{powered ankle-foot orthosis}
\newacronym{RMS}{RMS}{root-mean-square}
\newacronym{MOCAP}{MOCAP}{Motion Capture}
\newacronym{AP}{AP}{anteroposterior}
\title{\LARGE \bf
Influence of Motion Restrictions in an Ankle 
Exoskeleton on Gait Kinematics and Stability in Straight Walking
}
\author{Miha De\v{z}man, Charlotte Marquardt, Adnan {\"U}\u{g}{\"u}r and Tamim Asfour% <-this % stops a space
\thanks{\ackJuBot \newline \ackhht {\tt\small \{miha.dezman,asfour\}@kit.edu}}%
}

\begin{document}

\maketitle
\thispagestyle{fancy}
\pagestyle{fancy}
%\showpagenumbers % Displays page numbers -> disabled in final version of h2t_def!

%%%%%%%%%%%%%%%%%%%%%%%%%%%%%%%%%%%%%%%%%%%%%%%%%%%%%%%%%%%%%%%%%%%%%%%%%%%%%%%%
\begin{abstract}
Exoskeleton devices impose kinematic constraints on a user's motion and affect their stability due to added mass but also due to the simplified mechanical design.
This paper investigates how these constraints resulting from simplified mechanical designs impact the gait kinematics and stability of users by wearing an ankle exoskeleton with changeable degree of freedom (DoF). 
The exoskeleton used in this paper allows one, two, or three DoF at the ankle, simulating different levels of mechanical complexity.

This effect was evaluated in a pilot study consisting of six participants walking on a straight path.
The results show that increasing the exoskeleton DoF results in an improvement of several metrics, including kinematics and gait parameters.
The transition from 1 DoF to 2 DoF is shown to have a larger effect than the transition from 2 DoF to 3 DoF for an ankle exoskeleton.
However, an exoskeleton with 3 DoF at the ankle featured the best results.
Increasing the number of DoF resulted in stability values closer the values when walking without the exoskeleton, despite the added weight of the exoskeleton.
\end{abstract}

\section{Introduction}
%\the\textwidth
%\the\columnwidth

% ---- Why is the ankle joint and its DoF important during motion? ----
The ankle joint has three \acrfull{DoF} and can support loads up to four times the weight of the human body~\cite{kleipool2010relation}. 
It also plays a crucial role in generating positive power while walking~\cite{sawicki2008mechanics}. 
Exoskeleton devices that assist the ankle joint demonstrated significant decreases in metabolic energy expenditure~\cite{poggensee2021how} during support of straight walking, where ankle \gls{PFDF} movement is the most notable and the primary focus of the specific exoskeleton design.
However, exoskeleton designs should allow for all three rotations of the ankle, as the ankle undergoes movement in all three \gls{DoF} even during straight walking~\cite{hsu2008aaos}.
Furthermore, the ankle \gls{INEV} and \gls{IRER} \gls{DoF} are also more prominent when turning and walking on curved paths~\cite{Dezman2024Ankle}.
As a result, exoskeletons designed to solely support straight walking lack the versatility required for daily activities.

% ---- Mechanical cost of kinematic compatibility ----
The kinematic compatibility, \ie ability to adapt to the posture of a human joint, depends on the adaptability and \gls{DoF} of a specific design of the exoskeletons' frame and kinematics, as explored in our previous work~\cite{Dezman2024Ankle}.
The exoskeleton frame serves as the mechanical structure responsible for holding the exoskeleton components in place and transmitting the actuation torque to the cuffs.
Designing ankle exoskeletons with 3 \gls{DoF} involves a trade-off, and may quickly lead to increased mechanical complexity and weight in rigid designs, or a decrease in actuation forces in softsuit designs, as analysed in~\cite{Dezman2024Ankle}.

% ---- How does reduction of DoF affect the users motion? ----
The number of \gls{DoF}s of an ankle exoskeleton affects the motion and stability of the user, however, few studies have directly assessed this effect.
Choi~et~al.~\cite{choi2020effects} showed that a 2 \gls{DoF} \gls{PAFO} improved the user's stability significantly compared to a 1 \gls{DoF} \gls{PAFO}.
Using ski boots, Olivier~et~al.~\cite{olivier2015impact} investigated the effects of ankle restriction on hip and knee kinematics. 
The results showed more changes in hip kinematics compared to the knee joint.
Moreover, they observed that individuals adjust to the imposed restrictions with different strategies.
McCain~et~al.~\cite{mccain2021isolating} investigated the effect of restrictions on the ankle, knee and hip when using a 3D printed ankle stay and a knee brace to systematically limit the motion of these joints.
Restrictions revealed a detrimental effect on the metabolic expenditure of walking, reducing the peak ankle power and knee \gls{RoM}.
According to Ranaweera~et~al.~\cite{ranaweera2022effects}, restricting non-sagittal motions also affects muscle activation and causes significant changes in muscle activities.
These studies show that restrictions of \gls{DoF}s has a large effect on kinematics and stability.

% ---- How does added weight affect gait parameters, kinematics and metabolic cost? 
The literature put more emphasis on analyzing the influence of weight and inertia on gait, energetics and kinematics.
Heavy exoskeletons affect the \gls{RoM} of the ankle and knee and decrease the foot acceleration in the \gls{AP} direction~\cite{meuleman2013effect}.
Added inertia with \SI{3.5}{\kg} increases the swing times and affects anterior-posterior motion of both the pelvis and the head-arm-trunk segment~\cite{meuleman2013effect}.
Longer and slower strides are also observed by~\cite{browning2007effects}.
In addition, Jin~et~al.~\cite{jin2017effects} also reports reduction in step height and maximum knee flexion.
The placement of the supplementary weight is crucial, with distally placed weights intensifying the metabolic rates needed for leg swing, as reported by~\cite{browning2007effects}.
Longer strides reduce the \gls{AP} margin of stability, which is the ability to keep moving forward \cite{hak2013steps}.
The negative influences of supplementary weight and inertia are intertwined with the negative effects of the kinematic restrictions. 
Therefore, a thorough analysis should take into account both aspects.
  
% ---- Content & goal of this paper ----
The exoskeleton design requires a balance between mechanical complexity, strength and device weight. 
A lighter yet strong enough exoskeleton design may reduce the weight burden, however, it may also introduce limitations on kinematics and stability.
In this paper, we investigate the effects of the number of \gls{DoF} in an ankle exoskeleton on the user's gait, ankle kinematics and stability.
The participants walked with an exoskeleton that could simulate a one, two, or three \gls{DoF} at the ankle, thereby representing different levels of mechanical complexity.
We measured various gait parameters such as stride length, time, and height, as well as cuff rotation and the \gls{RoM}. 
Specifically, we analyzed the average values for \gls{PFDF}, \gls{INEV} and \gls{IRER}, their \gls{RoM} and similarity. 
Furthermore, our study investigates how these limitations affect the user's stability.

%-------------------------------------------------------------
The rest of the paper is organized as follows. 
\Cref{sec:Material_and_methods} describes the exoskeleton, user study, data postprocessing and analysis methods. 
\Cref{sec:results_and_analysis} reports and analyzes the results of the user study. 
\Cref{sec:Discussion} discusses the implications, limitations, and future work of this paper.
\Cref{sec:Conclusion} concludes the paper.

%%%%%%%%%%%%%%%%%%%%%%%%%%%%%%%%%%%%%%%%%%%%%%%%%%%%%%%%%%%%%%%%%%%%%%%%%%%%%%%%%%%%%%

\section{Materials and Methods}\label{sec:Material_and_methods}

This section introduces the design of the exoskeleton and the motion restriction method. 
It describes the conducted user study as well as the gait kinematic and stability metrics used to analyze the collected motion capture data.

\subsection{Exoskeleton and \gls{DoF} restriction}
\begin{figure}[h]
    \centering
    %\vspace{0.5em}
    %\def\svgwidth{0.95\linewidth}
    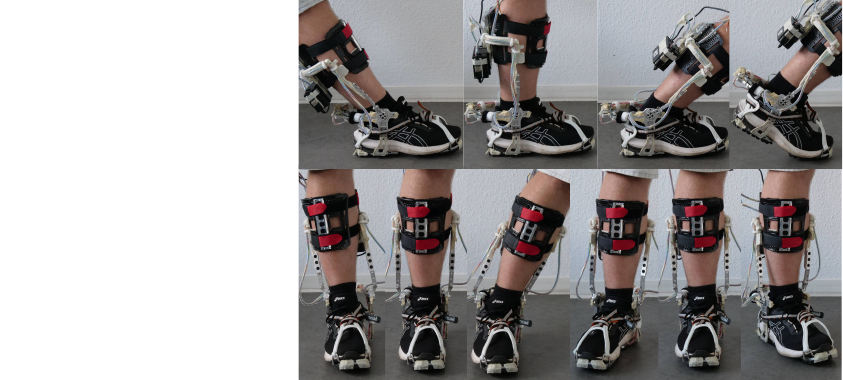
    %\vspace{-1.5em}
    \caption{\textbf{Left:} Shank and foot sections of the exoskeleton. 
    \textbf{Right:} The exoskeleton during the three rotations: \acrfull{PFDF}, \acrfull{INEV}, and \acrfull{IRER} of the ankle joint, as well as the forefoot rotation of the foot frame. (adapted from~\cite{Dezman2024Ankle})
        }
	\label{fig:Exo_3DoF}
    %\vspace{-1em}
\end{figure}

The ankle exoskeleton used in the study has been previously presented in~\cite{Dezman2024Ankle}.
It features a rigid frame design composed of a shank and foot section, shown in \cref{fig:Exo_3DoF} (left).
Both sections are joined through a parallelogram mechanism and multiple joints enabling  all three rotations of the ankle, as shown in \cref{fig:Exo_3DoF} (right).
An additional foot frame \gls{DoF} allows for forefoot rotation.

\begin{figure}[h]
    \centering
    \vspace{0.5em}
    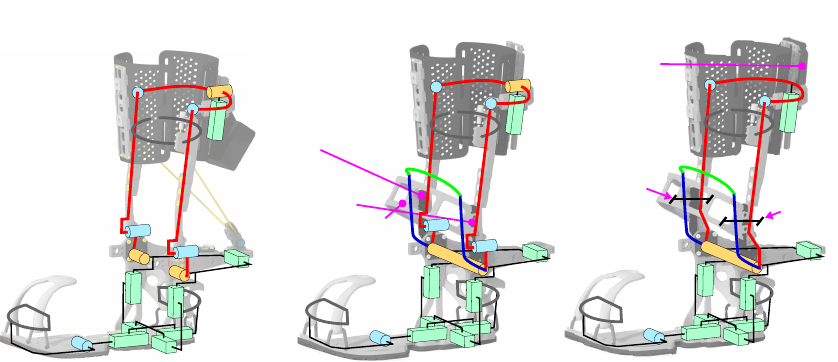
    %\vspace{-1.5em}
    \caption{
The three exoskeleton configurations: \emph{Exo3DoF}, \emph{Exo2DoF}, and \emph{Exo1DoF}. 
The kinematic structure comprises various joint types: a hinge joint represented by "\img{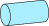}", a ball joint "\img{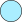}", a hinge joint with angle measurement "\img{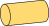}", and an adjustable translation joint "\img{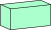}". The explanations for each callout are provided in the text. (adapted from~\cite{Dezman2024Ankle})
    }
    \label{fig:Exo_DoF_restriction}
    \vspace{-1.5em}
\end{figure}

\Cref{fig:Exo_DoF_restriction} shows the relevant kinematic joints for the 3 \gls{DoF} motion and size adjustment as well as the resulting three configurations of the exoskeleton \emph{Exo3DoF}, \emph{Exo2DoF}, and \emph{Exo1DoF}. 
The foot frame is adjustable in several dimensions, as denoted by adjustable translation joints in \cref{fig:Exo_DoF_restriction}.
This includes adjustment of length and width to accommodate different sizes and types of shoes.

To restrict the human ankle \gls{DoF} different exoskeleton frame joints must be immobilized. 
By mounting fixture elements, \ie blocking parts, on the existing exoskeleton structure, the ankle joint's \gls{DoF} can be reduced from three to either two or one resulting in the three configurations \emph{Exo3DoF}, \emph{Exo2DoF}, and \emph{Exo1DoF}. 
The following describes how the design restrictions were implemented.

The unrestricted 3 \gls{DoF} kinematics, \ie the \emph{Exo3DoF} case is shown in \Cref{fig:Exo_DoF_restriction} (left). 
It features three hinge joints with integrated angle measurement marked with \textcircled{{\scalebox{0.7}{a}}}, \textcircled{{\scalebox{0.7}{b}}} and \textcircled{{\scalebox{0.7}{g}}}, where \textcircled{{\scalebox{0.7}{a}}} and \textcircled{{\scalebox{0.7}{b}}} enable \gls{PFDF}.
The \gls{INEV} is enabled by hinge joints \textcircled{{\scalebox{0.7}{c}}}, \textcircled{{\scalebox{0.7}{d}}} and \textcircled{{\scalebox{0.7}{g}}}, which are ortogonal to the parallel rods and the axis of hinge joints {\textcircled{{\scalebox{0.7}{a}}}} and {\textcircled{{\scalebox{0.7}{b}}}}.
Additionally, the \gls{IRER} is enabled by ball joints \textcircled{{\scalebox{0.7}{e}}} and \textcircled{{\scalebox{0.7}{f}}}.

In the \emph{Exo2DoF} case, the exoskeleton mechanism restricts the \gls{IRER} by adding parts \textcircled{{\scalebox{0.7}{i}}}, \textcircled{{\scalebox{0.7}{j}}} and \textcircled{{\scalebox{0.7}{h}}}. 
These constrain the rotation of hinge joints \textcircled{{\scalebox{0.7}{a}}} and \textcircled{{\scalebox{0.7}{b}}}.
Consequently, the exoskeleton allows for \gls{PFDF} and \gls{INEV} of the ankle.

In the \emph{Exo1DoF} case, the \gls{INEV} is restricted by screwing parts {\textcircled{{\scalebox{0.7}{i}}}} and {\textcircled{{\scalebox{0.7}{j}}}} to the parallel rods, as denoted by {\textcircled{{\scalebox{0.7}{l}}}} and {\textcircled{{\scalebox{0.7}{m}}}}.
Furthermore, a fourth fixture part is added to constrain the hinge joint {\textcircled{{\scalebox{0.7}{g}}}}.
Consequently, only \gls{PFDF} motion is possible.

The \emph{ExoXDoF} configurations of \gls{DoF} restrictions, \ie \emph{Exo1DoF}, \emph{Exo2DoF} or \emph{Exo3DoF}, are chosen according to the most commonly reported \gls{DoF} combinations in ankle exoskeletons~\cite{aliman2017design}.
The entire ankle exoskeleton weighs \SI{1.8}{\kilo\gram}, wherein the foot frame section weighs \SI{0.65}{\kilo\gram}.
The fixture elements together weigh~\SI{100}{\gram}.
The exoskeleton is used passively; however, it has been designed to facilitate cable-driven actuation for plantar flexion motion in the future.

\subsection{User Study}
The goal of the user study is to assess how the exoskeleton \gls{DoF} constraints influence the kinematics, stability, and gait parameters of individuals while wearing the exoskeleton and walking straight.
Six healthy participants (four males and two females) took part in the study.
Their information is summarized in \cref{table:participant_information}.
\begin{table}[b]
\vspace{-0.5em}
\renewcommand{\arraystretch}{1.5}
\caption{Participant Information}
\label{table:participant_information}
%\vspace{-1em}
\begin{center}
    \begin{tabular}{|c|c|c|c|}
    \hline
    \textbf{Height {[}cm{]}} & \textbf{Weight {[}kg{]}} & \textbf{EU shoe size} & \textbf{Age {[}y{]}} \\
    \hline
    $177.7\pm\,9.3$ & $77.7\pm\,24.9$ & $42.7\pm\,2.5$ & $24.5\pm\,2.6$ \\
    \hline
\end{tabular}\\[1mm]
\scriptsize Values represent the mean and standard deviation.
\end{center}
\end{table}
%\vspace{-1em}
%
All participants provided written informed consent before the study participation and all methods were performed in accordance with the Declaration of Helsinki.
The experiment protocol was approved by the Karlsruhe Institute of Technology (KIT) Ethics Committee under ethical application for the JuBot project. 

Each participant walked with a self-selected speed along a \SI{3}{\m} straight path, turned around, walked back, and returned to the initial pose.
The walking path length was maximized while ensuring full motion capture functionality.
Each participant study session started with four repetitions of the \emph{NoExo} condition to establish a baseline measurement without the exoskeleton.
The three exoskeleton configurations followed, namely: \emph{Exo1DoF}, \emph{Exo2DoF} and \emph{Exo3DoF}, in a randomized order.
Same as for the \emph{NoExo} condition, each of the three configurations was repeated four times.
Each participant had a \SI{3}{\min} long familiarization phase before the session and resting pauses between conditions and repetitions.

The \emph{NoExo} case was performed without wearing of the exoskeleton and serves as a baseline.
In the configurations \emph{ExoXDoF}, the participants wore the exoskeleton on their right leg (\SI{1.8}{\kilo\gram}), where the exoskeleton allowed the \emph{X} number of \gls{DoF}.
The \gls{DoF} restriction approach was already explained in \cref{fig:Exo_DoF_restriction}.
The users also wore a foot frame on their left leg (\SI{0.65}{\kilo\gram}), to account for the thickness of the exoskeleton sole.
\begin{figure}[h]
    \centering
    %\vspace{0.5em}
    %\def\svgwidth{0.95\linewidth}
    %% Creator: Inkscape inkscape 0.92.5, www.inkscape.org
%% Accompanies image file 'user_study.pdf' (pdf, eps, ps)
\begingroup%
  \makeatletter%
  \providecommand\color[2][]{%
    \errmessage{(Inkscape) Color is used for the text in Inkscape, but the package 'color.sty' is not loaded}%
    \renewcommand\color[2][]{}%
  }%
  \providecommand\transparent[1]{%
    \errmessage{(Inkscape) Transparency is used (non-zero) for the text in Inkscape, but the package 'transparent.sty' is not loaded}%
    \renewcommand\transparent[1]{}%
  }%
  \providecommand\rotatebox[2]{#2}%
  \newcommand*\fsize{\dimexpr\f@size pt\relax}%
  \newcommand*\lineheight[1]{\fontsize{\fsize}{#1\fsize}\selectfont}%
  \ifx\svgwidth\undefined%
    \setlength{\unitlength}{238.84110819bp}%
    \ifx\svgscale\undefined%
      \relax%
    \else%
      \setlength{\unitlength}{\unitlength * \real{\svgscale}}%
    \fi%
  \else%
    \setlength{\unitlength}{\svgwidth}%
  \fi%
  \global\let\svgwidth\undefined%
  \global\let\svgscale\undefined%
  \makeatother%
  \begin{picture}(1,0.35605)%
    \lineheight{1}%
    \setlength\tabcolsep{0pt}%
    \put(0,0){\includegraphics[width=\unitlength,page=1]{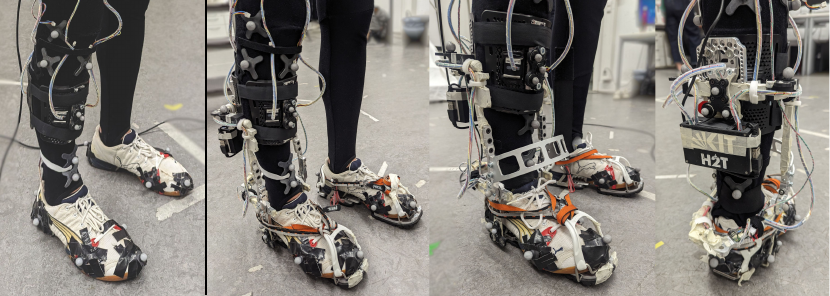}}%
    \put(0.00838742,0.32673779){\color[rgb]{1,1,1}\makebox(0,0)[lt]{\lineheight{1.25}\smash{\begin{tabular}[t]{l}\textbf{\textbf{NoExo}}\end{tabular}}}}%
    \put(0,0){\includegraphics[width=\unitlength,page=2]{user_study.pdf}}%
    \put(0.25663058,0.32612448){\color[rgb]{1,1,1}\makebox(0,0)[lt]{\lineheight{1.25}\smash{\begin{tabular}[t]{l}\textbf{\textbf{Exo3Dof}}\end{tabular}}}}%
    \put(0,0){\includegraphics[width=\unitlength,page=3]{user_study.pdf}}%
    \put(0.5257566,0.32612448){\color[rgb]{1,1,1}\makebox(0,0)[lt]{\lineheight{1.25}\smash{\begin{tabular}[t]{l}\textbf{\textbf{Exo2Dof}}\end{tabular}}}}%
    \put(0.79838039,0.32612448){\color[rgb]{1,1,1}\makebox(0,0)[lt]{\lineheight{1.25}\smash{\begin{tabular}[t]{l}\textbf{\textbf{Exo1Dof}}\end{tabular}}}}%
    \put(0,0){\includegraphics[width=\unitlength,page=4]{user_study.pdf}}%
  \end{picture}%
\endgroup%

    %\vspace{-1.5em}
    \caption{
    A participant without and with the ankle exoskeleton in its three configurations. 
    The parts added to fix certain \gls{DoF} are marked in red.
    }
    \label{fig:mocap_markers}
    %\vspace{-1em}
\end{figure}

\begin{figure}[t]
    \centering
    \vspace{0.25em}
    %\def\svgwidth{0.95\linewidth}
    %% Creator: Inkscape inkscape 0.92.5, www.inkscape.org
%% Accompanies image file 'Figure_Vicon_markers.pdf' (pdf, eps, ps)
\begingroup%
  \makeatletter%
  \providecommand\color[2][]{%
    \errmessage{(Inkscape) Color is used for the text in Inkscape, but the package 'color.sty' is not loaded}%
    \renewcommand\color[2][]{}%
  }%
  \providecommand\transparent[1]{%
    \errmessage{(Inkscape) Transparency is used (non-zero) for the text in Inkscape, but the package 'transparent.sty' is not loaded}%
    \renewcommand\transparent[1]{}%
  }%
  \providecommand\rotatebox[2]{#2}%
  \newcommand*\fsize{\dimexpr\f@size pt\relax}%
  \newcommand*\lineheight[1]{\fontsize{\fsize}{#1\fsize}\selectfont}%
  \ifx\svgwidth\undefined%
    \setlength{\unitlength}{182.76213158bp}%
    \ifx\svgscale\undefined%
      \relax%
    \else%
      \setlength{\unitlength}{\unitlength * \real{\svgscale}}%
    \fi%
  \else%
    \setlength{\unitlength}{\svgwidth}%
  \fi%
  \global\let\svgwidth\undefined%
  \global\let\svgscale\undefined%
  \makeatother%
  \begin{picture}(1,0.96732584)%
    \lineheight{1}%
    \setlength\tabcolsep{0pt}%
    \put(0,0){\includegraphics[width=\unitlength,page=1]{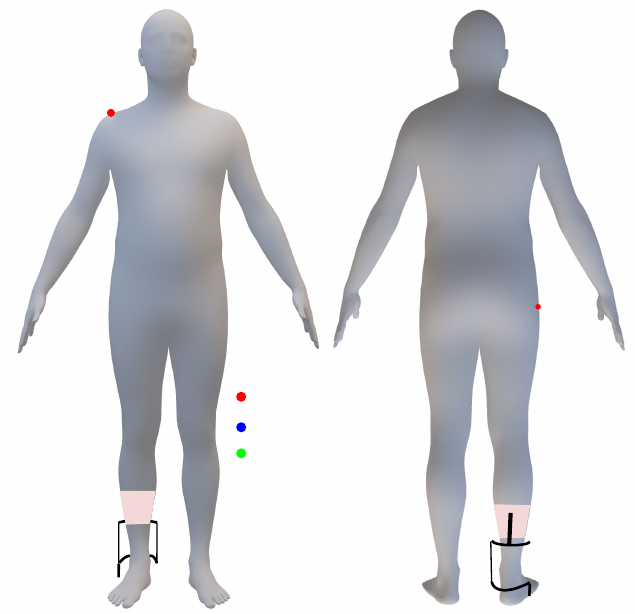}}%
    \put(0.40165432,0.32829329){\color[rgb]{0,0,0}\makebox(0,0)[lt]{\lineheight{1.25}\smash{\begin{tabular}[t]{l}user\end{tabular}}}}%
    \put(0.40117073,0.28025121){\color[rgb]{0,0,0}\makebox(0,0)[lt]{\lineheight{1.25}\smash{\begin{tabular}[t]{l}cuff\end{tabular}}}}%
    \put(0.40013682,0.2394467){\color[rgb]{0,0,0}\makebox(0,0)[lt]{\lineheight{1.25}\smash{\begin{tabular}[t]{l}exoskeleton\end{tabular}}}}%
    \put(0,0){\includegraphics[width=\unitlength,page=2]{Figure_Vicon_markers.pdf}}%
  \end{picture}%
\endgroup%

    %\vspace{-1.5em}
    \caption{
    \gls{MOCAP} marker positions on the exoskeleton and the user.
    Red markers are attached on the user. 
    Blue markers are attached on the exoskeleton cuff. 
    Green markers are attached on the exoskeleton.
    }
    \label{fig:mocap_marker_position}
    \vspace{-1em}
\end{figure}

The exoskeleton was adjusted during the donning process according to the visual inspection and the user's verbal comments regarding comfort and alignment.
A well-aligned exoskeleton features minimum vertical sliding of the shank cuff during \gls{PFDF}.
The motion of both the exoskeleton and the participant in the study was recorded by an optical \gls{MOCAP} system (Vicon Motion System, Ltd, UK). 
Passive markers were attached to the exoskeleton and participant in a way that ensured a continuous tracking of the markers needed to calculate all three rotations of the ankle joint in all configurations.
The marker configuration on the human body is shown in \cref{fig:mocap_markers} and \cref{fig:mocap_marker_position}. 
In addition to the \gls{MOCAP} data, the exoskeleton sensor data was also collected, including \gls{FMG} measurements, which are used in a different study that assesses the change of FMG signals due to the different restrictions in the exoskeleton conditions.
%analyzed and presented in a different paper~\cite{Marquardt2024}.

\subsection{Stride Segmentation}

\begin{figure}[b]
    \centering
    %\vspace{0.5em}
    %\def\svgwidth{0.95\linewidth}
    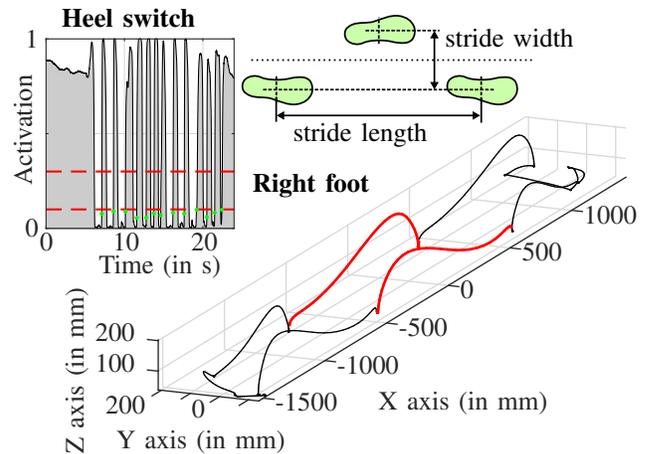
    %\vspace{-1.5em}
    \caption{
    \textbf{Top left:} heel switch activation and detected strides in green.
    \textbf{Bottom:} an example of two segmented strides of the left leg.
    \textbf{Top right:} the stride length and width definition.
    }
    \label{fig:Exo_stride_segmentation}
    %\vspace{-1.5em}
\end{figure}

All parameters and signals were evaluated on a stride basis, therefore all raw measurements were segmented into strides and averaged.
The angles were segmented using the heel switch activation as seen in \cref{fig:Exo_stride_segmentation} (top left). 
All transient steps were removed to avoid starting and ending transitions.
The selection criterion was the crossing of the zero absolute position of the X axis, as seen in \cref{fig:Exo_stride_segmentation} (bottom).

\subsection{Gait Parameters}

The analyzed gait parameters used to assess the effect the \gls{DoF} reduction has on participant gait include:
\begin{enumerate}
    \item \emph{stride length}, analyzed for both legs, denoting the distance from one heel strike to the next heel strike (shown in \cref{fig:Exo_stride_segmentation} (top right)). The stride length is analysed for each leg independently.
    \item \emph{stride width} denotes the maximum distance between both legs (shown in \cref{fig:Exo_stride_segmentation} (top right)).
    \item \emph{stride time} denotes the time between two heel strikes.
    \item \emph{foot height} denotes the maximum height of the foot during a stride for each leg independently.
\end{enumerate}

\subsection{Kinematics}

The effect on a user's kinematics is determined by observing the three ankle motions: \gls{PFDF}, \gls{INEV}, and \gls{IRER}.
For each angle, the averages and standard deviations of all strides of all participants were calculated for each exoskeleton configuration.
The \gls{RMSE} was used to compare the joint angle trajectories of \emph{ExoXDoF} against the \emph{NoExo} condition. 
A lower \gls{RMSE} indicates a higher degree of similarity between two conditions. 
The \gls{RoM} denotes the maximum and minimum values of the trajectory average and the maximum and minimum standard deviation reached during the gait cycle.

Additionally, the rotation of the shank cuff around the shank axis is measured using markers placed on the exoskeleton and on the user.
The shank cuff rotation indicates the angle between the cuff and the knee joint.
An increased cuff rotation indicates more movement of the knee relative the shank cuff.
This means the presence of \gls{IRER} motion despite the exoskeleton restriction, due to the compliance of the soft tissues around the shank.

\subsection{Stability}

The participant stability is evaluated using a stabilogram of the trunk roll and pitch acceleration, as outlined in~\cite{choi2020effects}.
Maintaining balance of the upper body is important to avoid falling. 
Therefore, the stability is assessed based on the magnitudes of the detected accelerations of the upper body.
To quantify them, a Gaussian ellipsoid is fitted onto the accelerations, and its two eigenvalue vectors are used to calculate a \gls{RMS} value representing the instability.
According to~\cite{choi2020effects}, a higher \gls{RMS} value, \ie the instability value, denotes greater swing of the trunk and indicates lower stability of the user. 
In the current study, trunk accelerations are calculated based on marker motion, positioned in the lower back and between the shoulder girdle.
However, an \gls{IMU} based system may also be used~\cite{choi2020effects}.

\section{Results and Analysis}\label{sec:results_and_analysis}

\subsection{Angle and \gls{RoM}}
\begin{figure*}[b]
    \centering
    \vspace{0.5em}
    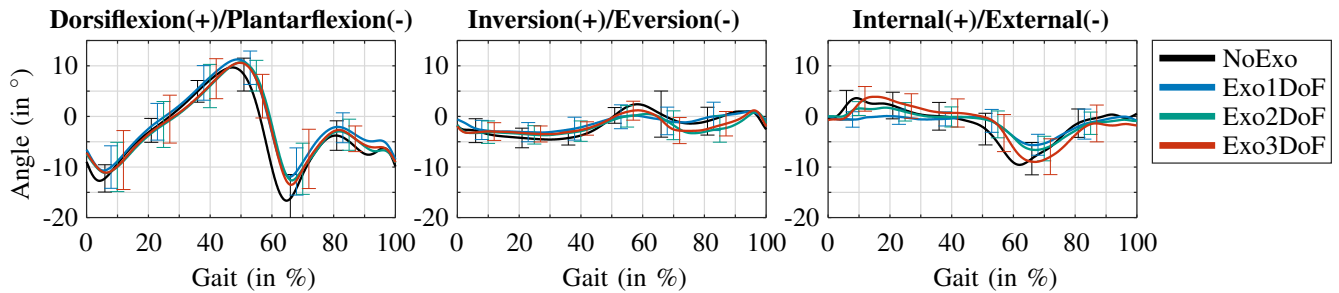
    %\includegraphics[width=8cm]{figures/Plot_Ankle_Angle.pdf}
    %\vspace{-1.5em}
    \caption{Average of ankle \gls{PFDF}, \gls{INEV} and \gls{IRER} of all strides shown for each configuration. 
    The motion direction is indicated in the title of the subfigures. 
    Error bars indicating the standard deviation of all strides are also given.
    }\label{fig:Ankle_Angles}
    \vspace{-1em}
\end{figure*}
This section presents angle measurements and \gls{RoM} results. 
\Cref{fig:Ankle_Angles} shows the averages and standard deviations of all strides for the three ankle rotations (\gls{PFDF}, \gls{INEV} and \gls{IRER}) and all configurations.
The averaged curves maintained a similar in shape in all configurations compared to the \emph{NoExo} condition.
However, some discrepancies were observed.
A slight phase delay was observed for \gls{PFDF} and \gls{IRER} for all exoskeleton conditions compared to the \emph{NoExo} condition.
The standard deviation of \emph{NoExo} condition is generally small but increases at push-off (\SI{60}{\percent}), especially for \gls{PFDF} and \gls{INEV}.
The exoskeleton configurations feature a similar standard deviation for all strides for \gls{INEV} and \gls{IRER} compared to the \emph{NoExo} condition.
However, the standard deviation is higher for \gls{PFDF}.
The most prominent differences between the shape of all curves are evident in \gls{IRER}.
In this case, the \emph{Exo1DoF} curve is flat in the first \SIrange{0}{40}{\percent} of gait cycle.
Changing the exoskeleton to the 2 \gls{DoF} configuration results in an increase in angle magnitudes, but only \emph{Exo3DoF} returns its shape closer to the shape of the \emph{NoExo} curve.

\begin{figure}[h]
    \centering
    %\vspace{0.5em}
    %\def\svgwidth{0.95\linewidth}
    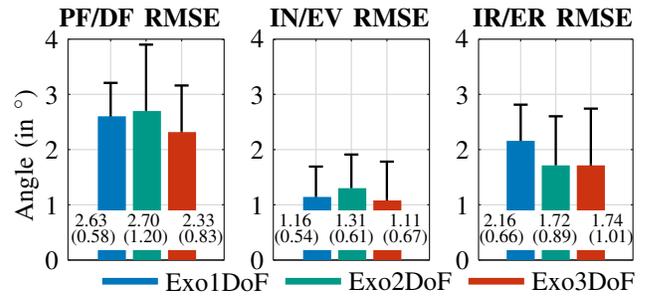
    %\includegraphics[width=8cm]{figures/Plot_Ankle_Angle.pdf}
    %\vspace{-1.5em}
    \caption{ Ankle rotation similarity of \emph{ExoXDoF} to \emph{NoExo} condition.
    The \gls{RMSE} values are calculated between the averages of the respective configuration and the \emph{NoExo} average for all strides of all participants.
    The standard deviation (vertical error lines) of \gls{RMSE} values shows variability between the participants.
    }
    \label{fig:Ankle_Angles_RMSE}
    %\vspace{-0.5em}
\end{figure}

The similarity between curves is better depicted in \cref{fig:Ankle_Angles_RMSE}.
For \gls{PFDF}, \emph{Exo3DoF} is the most similar to the \emph{NoExo} case, as depicted by the smallest \gls{RMSE} value.
For \gls{INEV}, \emph{Exo3DoF} features the smallest \gls{RMSE} value, however, the similarity is similar to the \emph{Exo1DoF} case.
For \gls{IRER}, the values of \emph{Exo3DoF} and \emph{Exo2DoF} are similar, but lower than \emph{Exo1DoF} case.

\begin{figure}[ht]
    \centering
    \vspace{0.5em}
    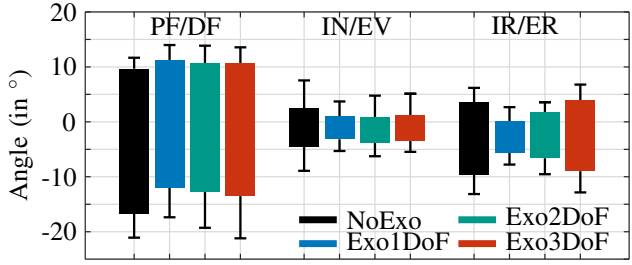
    %\includegraphics[width=8cm]{figures/Plot_Ankle_Angle.pdf}
    %\vspace{-1.5em}
    \caption{ 
    \gls{RoM} for all three ankle rotations during straight walking for all configurations (\emph{NoExo}, \emph{Exo1DoF}, \emph{Exo2DoF} and \emph{Exo3DoF}).
    The blocks represent the maximum and the mininum angle of \cref{fig:Ankle_Angles}.
    The error ticks represent the maximum and minimum when standard deviation is added/subtracted from the curve.
    }
    \label{fig:Ankle_RoM}
    \vspace{-1em}
\end{figure}

Finally, \cref{fig:Ankle_RoM} shows the \gls{RoM} of the average curves of \cref{fig:Ankle_Angles}.
The \gls{PFDF} \gls{RoM} shifts with the introduction of the exoskeleton but appears to shift closer to the \gls{RoM} of the \emph{NoExo} case with the increasing \gls{DoF}.
\gls{INEV} \gls{RoM} shows only a small improvement with the increase of \gls{DoF}.
\gls{IRER} \gls{RoM} features the highest improvement of \gls{RoM} with increasing \gls{DoF} and returns it to the \gls{RoM} of the \emph{NoExo} case.

\subsection{Gait parameters}
\begin{figure}[h]
    \centering
    %\vspace{0.5em}
    %\def\svgwidth{0.95\linewidth}
    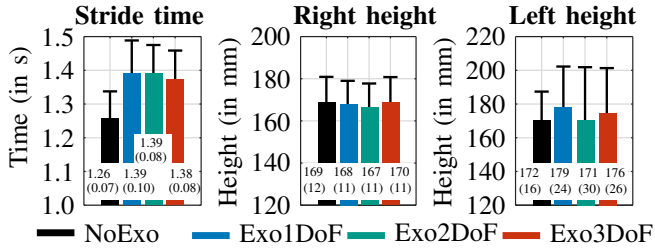
    %\includegraphics[width=8cm]{figures/Plot_Ankle_Angle.pdf}
    %\vspace{-1.5em}
    \caption{ 
    The left graph displays the average stride duration of all participants for each condition. 
    The middle graph illustrates the maximum height reached by the right foot during a stride, while the right graph shows the same for the left leg. 
    The vertical error lines represent the standard deviation, indicating the variation between participants.
    }
	\label{fig:Stride_time_foot_height}
    %\vspace{-1em}
\end{figure}

This section presents the resulting gait parameters: stride time, foot height, stride length, and width.
The stride time and foot height are shown in \cref{fig:Stride_time_foot_height}.
Wearing of the exoskeleton increases the stride time with slight time improvement as the number of \gls{DoF} increases towards the \emph{NoExo} condition.
%The standard deviation remains nearly the same over all conditions.
The average and standard deviation of the height of the right foot remain nearly the same throughout all configurations.
However, the average and standard deviation of the left foot height fluctuate when changing between configurations.

\begin{figure}[h]
    \centering
    \vspace{0.5em}
    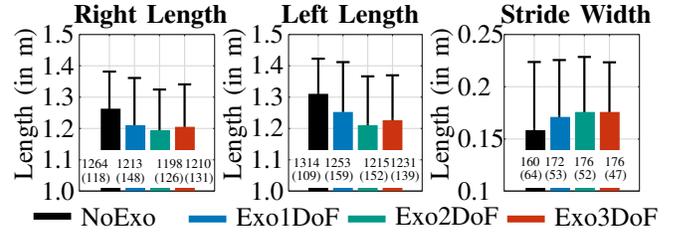
    %\includegraphics[width=8cm]{figures/Plot_Ankle_Angle.pdf}
    %\vspace{-1.5em}
    \caption{  
    The left graph shows the average time and standard deviation of all the strides of all participants shown for each condition.
    The middle and right graphs show the average stride length and standard deviation for the left and right legs, respectively, across various configurations, for all strides of all participants.%\remark{Check caption!}
    }
    \label{fig:Stride_length_width}
    \vspace{-1em}
\end{figure}

The stride length for both feet and the stride width for each condition are shown in \cref{fig:Stride_length_width}.
The right leg stride length, \ie the side wearing the exoskeleton, shows decreased step length for all \emph{ExoXDoF} configurations.
Changing the number of \gls{DoF}s did not noticeably affect it.
Left stride length decreases for \emph{ExoXDoF} configurations.
Both the left and right strides show a comparable standard deviation that does not change during the conditions.
The stride width increases with the increase of \gls{DoF}, in addition, the standard deviation becomes smaller.

\begin{figure}[ht]
    \centering
    %\vspace{0.5em}
    %\def\svgwidth{0.95\linewidth}
    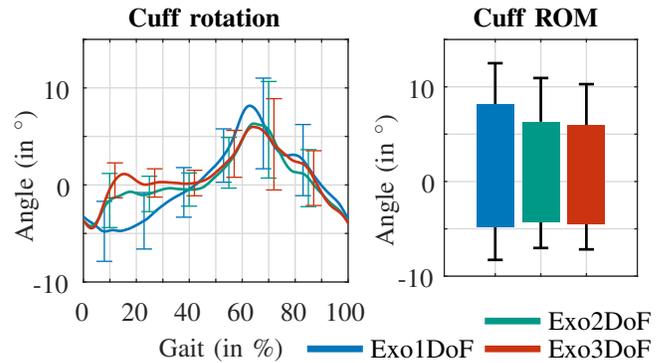
    %\includegraphics[width=8cm]{figures/Plot_Ankle_Angle.pdf}
    %\vspace{-1.5em}
    \caption{ 
    The left graph shows the average cuff rotation for all strides of all participants shown for each condition. 
    The error bars represent the standard deviation of all strides.
    The blocks represent the maximum and minimum rotation of the cuff shown in the left graph.
    The error ticks represent the maximum and minimum when standard deviation is added/subtracted from the curve.
    }
	\label{fig:Cuff_rotation}
    %\vspace{-1.5em}
\end{figure}

The relative cuff rotation displayed in \cref{fig:Cuff_rotation} (left) offers a deeper look on the behaviour of \gls{IRER} between different configurations.
The cuff rotation is the largest for the \emph{Exo1DoF} case.
With increasing the number of \gls{DoF}, the relative rotation of the cuff starts to decrease, indicating smaller rotations of the cuff relative to the knee. 
Condition \emph{Exo3DoF} displays the smallest rotation.

\Cref{fig:Cuff_rotation} (right) presents the \gls{RoM} values.
Changing from \emph{Exo1DoF} to \emph{Exo2DoF} shows a noticeable decrease in cuff \gls{RoM}.
However, changing from \emph{Exo2DoF} to \emph{Exo3DoF} does not show a noticeable improvement in cuff \gls{RoM}.

%Continue here ---------------------------------------------------------------------------------------------------------------------------------------------------

\subsection{Stability evaluation}
\begin{figure*}[ht]
    \centering
    %\vspace{0.5em}
    %\def\svgwidth{0.95\linewidth}
    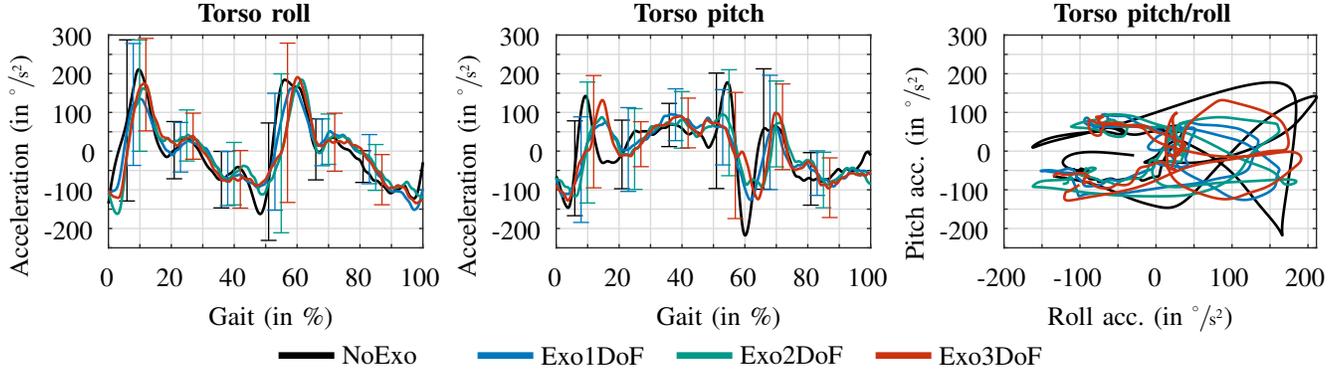
    %\includegraphics[width=8cm]{figures/Plot_Ankle_Angle.pdf}
    %\vspace{-1.5em}
    \caption{ 
    The left graph shows the average torso roll acceleration for all the strides of all participants shown for all conditions. 
    The error bars represent the standard deviation of all strides.
    In the same way, the middle graph shows the average acceleration of the torso pitch with the standard deviation.
    The right graph shows the correlation of both accelerations in a 2D representation.
    }
    \label{fig:Torso_roll_pitch}
    \vspace{-1.5em}
\end{figure*}

\Cref{fig:Torso_roll_pitch} (left, middle) displays the average roll and pitch accelerations of the trunk.
The \emph{NoExo} condition features the highest accelerations.
However, wearing the exoskeleton (\emph{ExoXDoF} condition) reduced some of the acceleration peaks compared to \emph{NoExo} condition.
Furthermore, several acceleration peaks of the \emph{NoExo} condition appear to precede those of the \emph{ExoXDoF} condition.

\begin{figure*}[hb]
    \centering
    \vspace{-0.5em}
    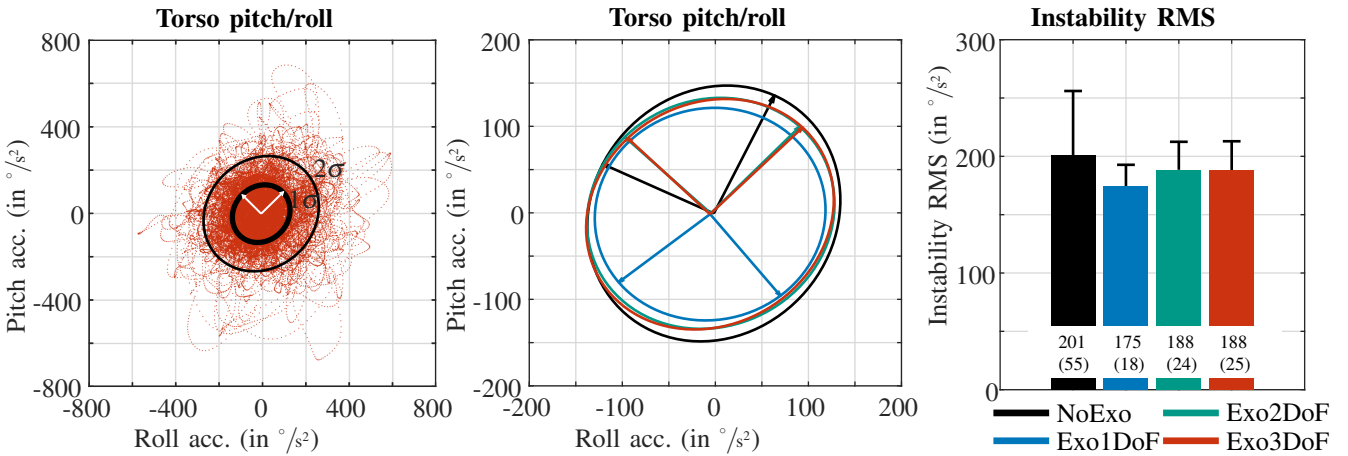
    %\vspace{-1.5em}
    \caption{ 
        The left graph shows the torso acceleration data for roll and pitch rotation and the corresponding Gaussian ellipsoid fit for the \emph{Exo3DoF} configuration.
        The middle graph displays the Gaussian ellipsoids for all configurations.
        The right graph presents the mean instability value for each configuration, averaged over all participants.
        The error bars indicate the standard deviation for the averages of all participants.}
	\label{fig:Stabilogram_Stability}
    %\vspace{-1.5em}
\end{figure*}

The stabilograms and instability shown in \cref{fig:Stabilogram_Stability} are based on data from \cref{fig:Torso_roll_pitch}.
\Cref{fig:Stabilogram_Stability} (left) shows an example Gaussian ellipsoid fit ($1\sigma$ and $2\sigma$) for the Pitch/Roll accelerations for the \emph{Exo3DoF} case with the corresponding eigenvectors.
\Cref{fig:Stabilogram_Stability} (middle) shows the ellipsoids ($1\sigma$) and the eigenvectors for all configurations.
\Cref{fig:Stabilogram_Stability} (right) shows the resulting instability based on the \gls{RMS} of the corresponding eigenvectors shown for each configuration.
Wearing of the exoskeleton (configurations \emph{ExoXDoF}) results in a smaller ellipsoid (see \cref{fig:Stabilogram_Stability} (middle)) and consequently lower instability values.
The ellipsoid becomes larger, especially wider in roll axis, by releasing the \gls{DoF} restriction from 1 \gls{DoF} to either 2 \gls{DoF} or 3 \gls{DoF}.
The change from 1 \gls{DoF} to 2 \gls{DoF} is greater than the one from 2 \gls{DoF} to 3 \gls{DoF}.
\Cref{fig:Stabilogram_Stability} (right) shows the highest variability (standard deviation) between participants for the \emph{NoExo} case.
However, wearing the exoskeleton (\emph{ExoXDoF}) reduces the standard deviation, indicating that it introduces some restrictions that are similar between participants.
Removing the exoskeleton restrictions \gls{DoF}, \ie moving to the configuration \emph{Exo3DoF}, enhance instability values but does not restore them to the \emph{NoExo} values.

\section{Discussion}\label{sec:Discussion}

The study shows that wearing the exoskeleton results in a noticeable impact on the user's kinematics, gait parameters and stability.
Increasing the number of \gls{DoF} improved the kinematic compatibility of the exoskeleton, demonstrated by a larger \gls{RoM} of \gls{PFDF} and \gls{IRER}.
However, \gls{INEV} does not show the same level of improvement.
Increased \gls{DoF} improves stride time, reduces cuff rotation relative to the knee, and increases the instability value.
However, the instability values in~\cite{choi2020effects} demonstrated a different effect showing a decrease when switching from the 1 \gls{DoF} to 2 \gls{DoF} configuration.
Our study demonstrated the opposite behavior, as the instability value increased with more \gls{DoF}, resulting in reduced stability of the participant.
Based on our findings, the exoskeleton device discussed in~\cite{choi2020effects} induces greater restrictions on users while wearing the 2 \gls{DoF} exoskeleton in an unpowered case.
While \cite{choi2020effects} does not provide stability values without the exoskeleton, additional investigation is required to validate these findings.

The criteria related to \gls{IRER}, that is \gls{RMSE} value from \cref{fig:Ankle_Angles_RMSE}, cuff rotation values from \cref{fig:Cuff_rotation}, and the stability values in \cref{fig:Stabilogram_Stability} (right), revealed larger improvements when transitioning from 1 \gls{DoF} to 2 \gls{DoF} than when transitioning from 2 \gls{DoF} to 3 \gls{DoF}.
%Such observation was also confirmed in the \gls{FMG} measurement part of our study presented in~\cite{Marquardt2024}.
In contrast, restriction of the exoskeleton \gls{IRER} has a limited influence on the \gls{IRER} rotation of the ankle, since rotation is still possible due to the soft tissues around the shank.
This is shown in \cref{fig:Cuff_rotation} (right), as even under \gls{DoF} restrictions the ankle can move, especially in \gls{IRER}.
It is expected that the \gls{IRER} becomes more relevant when an ankle exoskeleton is combined with a knee exoskeleton.

This work has limitations related to both the exoskeleton and study design and execution.
The exoskeleton's weight (\SI{1.8}{\kilo\gram}) is the first hardware limitation, negatively affecting the gait.
Previous studies have shown that adding weight loads on the foot increases stride length and time, and decreases step height \cite{jin2017effects, browning2007effects}.
Our study demonstrated longer stride times while wearing the exoskeleton and a shorter stride length. 
However, foot height remained similar between conditions. 
Therefore, the added weight does not have the same influence as reported in previous studies. 
A possible explanation is that the exoskeleton prompted users to adapt a more cautious walking style, leading to slower and shorter strides.
The short walking path also played a role in this, as it prevented the users from achieving a stable gait.
A longer path would enable the users to adapt to the exoskeleton and walk more naturally, once they reached a stable gait.

The rigid sole is another hardware limitation that negatively affects the selected criteria.
Previous studies have shown that rigid-soled shoes cause short phase delays in the average curves~\cite{schmitthenner2020effect}.
We observed the same phenomenon in our study, as shown in \cref{fig:Ankle_Angles}.
Therefore, the removal of the rigid sole is identified as an important next step in the development of this exoskeleton.
Moreover, the exoskeleton's adjustments were insufficient for one of the six participants. 
Therefore, the next exoskeleton iteration will also feature a larger manual adjustment range.

\section{Conclusion}\label{sec:Conclusion}

This paper addressed the impact of \gls{DoF} reductions on users and the role of ankle \gls{DoF} in restoring ankle kinematics, gait parameters, and stability for straight walking.
The findings demonstrated that enabling of \gls{INEV} on the exoskeleton yielded greater improvements compared to subsequently adding \gls{IRER} rotation.
However, the greatest improvement was observed when the ankle exoskeleton allowed for all three \gls{DoF}. 
It is expected that \gls{IRER} becomes more important when the ankle exoskeleton includes a knee joint.
This implies that exoskeletons featuring solely \gls{PFDF} might benefit from extensions to the other \gls{DoF}.

For future work, we are developing a second iteration of the ankle exoskeleton with a knee joint extension. 
The device will have a lighter frame and a flexible sole. 
These improvements will enable further investigations of the findings of this study.

%%%%%%%%%%%%%%%%%%%%%%%%%%%%%%%%%%%%%%%%%%%%%%%%%%%%%%%%%%%%%%%%%%%%%%%%%%%%%%%%
%\vspace{-1em}
%\section*{ACKNOWLEDGMENT}
%\ackJuBot

%%%%%%%%%%%%%%%%%%%%%%%%%%%%%%%%%%%%%%%%%%%%%%%%%%%%%%%%%%%%%%%%%%%%%%%%%%%%%%%%

%\clearpage
\bibliographystyle{IEEEtran}
\bibliography{IEEEabrv,HumanoidsGroup,Ankle_Exoskeleton}

\end{document}